\documentclass[conference]{IEEEtran}
\IEEEoverridecommandlockouts

\usepackage{cite}
\usepackage{amsmath,amssymb,amsfonts}
\usepackage{algorithmic}
\usepackage{graphicx}
\usepackage{textcomp}
\usepackage{xcolor}
\def\BibTeX{{\rm B\kern-.05em{\sc i\kern-.025em b}\kern-.08em
    T\kern-.1667em\lower.7ex\hbox{E}\kern-.125emX}}
\begin{document}

\title{A Hybrid Attention Framework for Fake News Detection with Large Language Models
	\thanks{$^\ast$ Corresponding author: xiaochux@alumni.cmu.edu}
	\thanks{$^1$ These authors contributed equally to this work.}
}

\author{
\IEEEauthorblockN{Xiaochuan Xu\textsuperscript{$\ast$1}}
\IEEEauthorblockA{\textit{Information Networking Institute} \\
\textit{Carnegie Mellon University}\\
Pittsburgh, USA \\
xiaochux@alumni.cmu.edu} \\
\IEEEauthorblockN{Zeqiu Xu\textsuperscript{2}}
\IEEEauthorblockA{\textit{Information Networking Institute} \\
	\textit{Carnegie Mellon University}\\
	Pittsburgh, USA \\
	zeqiux@alumni.cmu.edu}
\and 
\IEEEauthorblockN{Peiyang Yu\textsuperscript{1}}
\IEEEauthorblockA{\textit{Information Networking Institute} \\
\textit{Carnegie Mellon University}\\
Pittsburgh, USA \\
peiyangy@alumni.cmu.edu} \\
\IEEEauthorblockN{Jiani Wang\textsuperscript{3}}
\IEEEauthorblockA{\textit{Department of Computer Science} \\
\textit{Stanford University}\\
Stanford, USA \\
jianiw@alumni.stanford.edu}
}

\setlength{\baselineskip}{11pt} 

\maketitle

\begin{abstract}
With the rapid growth of online information, the spread of fake news has become a serious social challenge. In this study, we propose a novel detection framework based on Large Language Models (LLMs) to identify and classify fake news by integrating textual statistical features and deep semantic features. Our approach utilizes the contextual understanding capability of the large language model for text analysis and introduces a hybrid attention mechanism to focus on feature combinations that are particularly important for fake news identification. Extensive experiments on the WELFake news dataset show that our model significantly outperforms existing methods, with a 1.5\% improvement in F1 score. In addition, we assess the interpretability of the model through attention heat maps and SHAP values, providing actionable insights for content review strategies. Our framework provides a scalable and efficient solution to deal with the spread of fake news and helps build a more reliable online information ecosystem.
\end{abstract}

\begin{IEEEkeywords}
Fake news detection, large language modeling, feature fusion, content review, online security, interpretability, attention mechanism.
\end{IEEEkeywords}

\section{Introduction}

\IEEEPARstart{I}{n} the digital information age, the rapid spread of fake news has become a serious global challenge. Studies have shown that fake news spreads significantly faster than real news on social media platforms, and its reach may reach millions of users in a matter of hours~\cite{1}. This phenomenon not only threatens the public's right of access to truthful information, but also may lead to social cognitive divisions, economic losses, and even public safety crises.

Traditional news authentication relies mainly on manual verification and professional fact-checking organizations, a method that seems inadequate in the face of massive online information. Especially during emergencies and major social events, fake news is often widely disseminated before it can be verified. This situation highlights the urgency of developing automated and efficient fake news detection systems~\cite{2}.

In recent years, thanks to the development of large language models (LLMs), automated fake news detection has made significant progress~\cite{3}. However, existing methods mainly focus on the semantic analysis of text content, ignoring other important features of news texts, such as headline features, text structure features, and sentiment tendencies. Our study found that fake news often exhibits unique linguistic patterns, for example, features including the proportion of capital letters in headlines, the frequency of punctuation use, and the frequency of numerical occurrences, which may be important indicators for identifying fake news. 

To address this issue, this study proposes a comprehensive fake news detection framework based on the WELFake News dataset. The framework not only considers the semantic content of the news text, but also integrates multidimensional features including text length, punctuation distribution, capitalization ratio, numerical frequency, and sentiment polarity. Through the hybrid attention mechanism, our model is able to automatically identify and focus on these feature combinations that are particularly important for fake news recognition. Experimental results show that our method achieves significant improvements in accuracy and robustness, with a 1.5\% increase in F1 score, compared to traditional methods that rely only on text content analysis~\cite{4}.

In addition, we pay special attention to the issue of model interpretability. By analyzing the attention heat map and SHAP values, we are able to clearly demonstrate the specific features that the model relies on when making judgments, which not only improves the transparency of the model's decision making, but also provides a practical reference index for journalists and content reviewers~\cite{1}. Our study provides a new technical path for building a more reliable online information ecosystem, and also provides important practical experience and theoretical basis for subsequent research on fake news detection.

\section{Methodology}

\begin{figure}[!t]
	\centering
	\includegraphics[width=\linewidth]{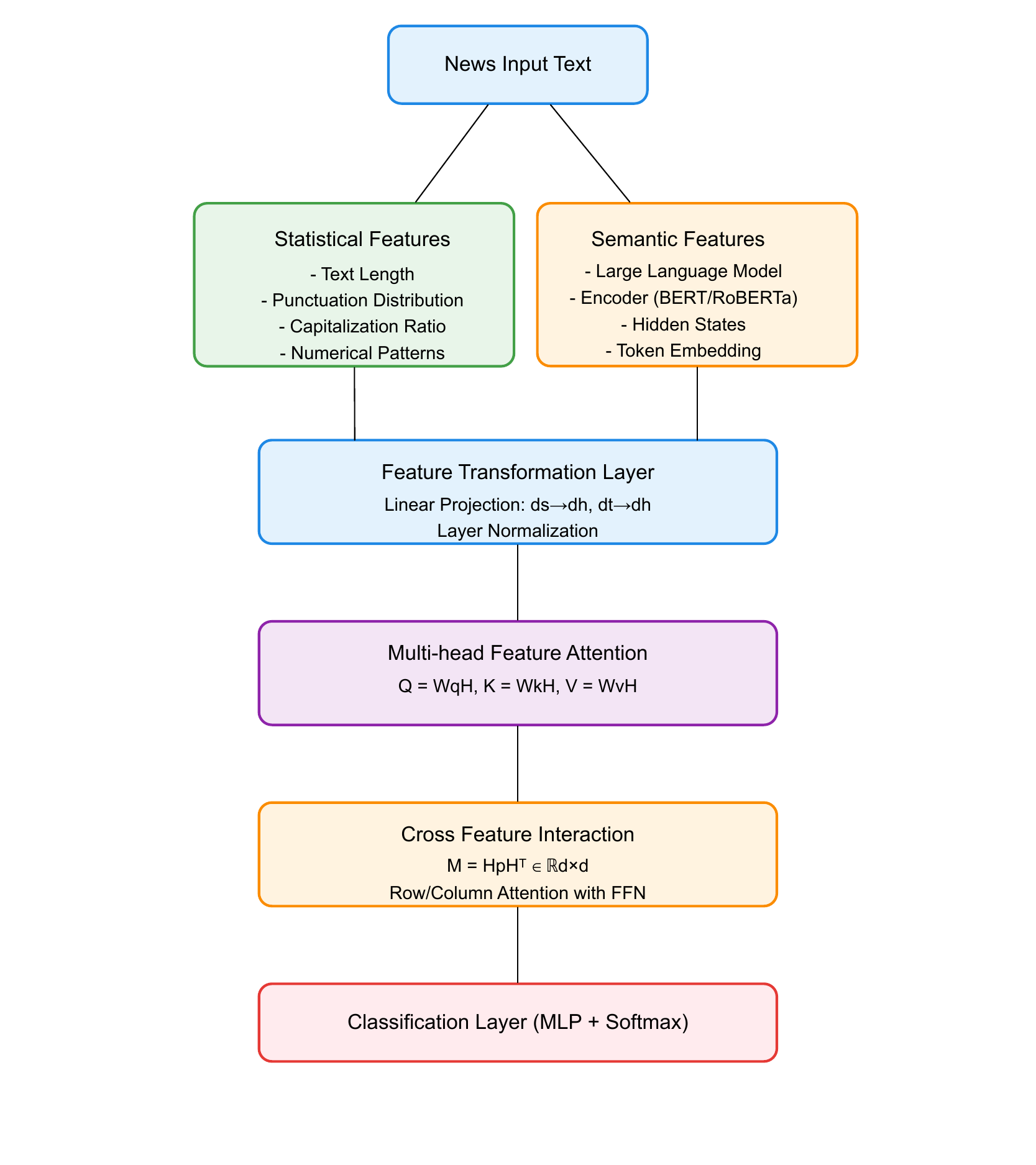}
	\caption{Model architecture diagram.}
	\label{fig:1}
\end{figure}

As shown in Fig.~\ref{fig:1}, this study proposes a fake news detection framework based on hybrid feature fusion. The framework achieves efficient detection of fake news by integrating statistical and semantic features of text and introducing a two-layer attention mechanism for feature fusion and interaction. Fig.~\ref{fig:1} shows the overall architecture of the model, which mainly contains four key modules: feature extraction, feature transformation, multi-head feature attention and cross-feature interaction. The core innovation of the model lies in the design of a hierarchical attention mechanism to capture the complex relationships between different types of features, in which the computation of feature attention scores can be expressed as:
\begin{align}
	\alpha_{ij}&=\frac{\exp (e_{ij})}{\sum\limits_{k=1}^n \exp (e_{ik})}, \\
	e_{ij}&=\frac{(W_q h_i )^T (W_k h_j)}{\sqrt{d_k}}
\end{align}

\noindent where $h_i$ and $h_j$ denote the input feature vectors, $W_q$ and $W_k$ are the learnable parameter matrices, and $d_k$ is the feature dimension, respectively.

\subsection{Feature Extraction}

In the feature extraction stage, the model extracts features from news text in two dimensions. The statistical feature part focuses on the formal features of the text, including text length, punctuation distribution, capital letter proportion and numerical value occurrence pattern, etc. These features can effectively capture the uniqueness of fake news in terms of writing style. The semantic feature part, on the other hand, utilizes a pre-trained language model (BERT/RoBERTa) to encode the text and obtain the deep semantic representation. Specifically, we extract the [CLS] tag embedding in the last layer of the model as the semantic feature representation of the text, which contains the global semantic information of the whole text~\cite{5}.

\subsection{Feature Transformation and Attention Mechanism}

In order to realize the effective fusion of features, the model first maps statistical features and semantic features to the same feature space through a feature transformation layer. The feature transformation layer contains linear projection and layer normalization operations to convert the feature vectors with dimensions ds and dt, respectively, into a representation with uniform dimension dh. Based on this, a multi-head feature attention mechanism is introduced to learn the association between features~\cite{6}. This mechanism achieves dynamic weighting of features through query-key-value (Q-K-V) attention computation, which enables the model to adaptively focus on important feature combinations.

\subsection{Cross-feature Interaction and Classification}

After obtaining the attention-weighted feature representation, the model further enhances the information interaction between features through the cross-feature interaction layer. This layer constructs the feature interaction matrix M, which captures the dependencies between different types of features through the ranks attention mechanism. The dimension of the feature interaction matrix is dh×dh, and its computation considers the contributions of both statistical and semantic features. Finally, the features processed by the feed-forward neural network (FFN) are fed into the classification layer for the final authenticity judgment by means of the multilayer perceptron (MLP) and Softmax function. This hierarchical feature processing allows the model to make full use of the complementary information between different types of features to improve the accuracy of detection~\cite{7}.

\section{Experiments}

\subsection{Data Preprocessing}

The enhanced WELFake news dataset is used as the experimental data in this study. This dataset is an optimized version of the original WELFake dataset, which contains 62,308 news samples, in which the ratio of real news and fake news is close to 1:1, ensuring the basic balance of the data. In the data preprocessing stage, we used natural language processing tools such as NLTK and spaCy to systematically clean and standardize the text, including basic operations such as lowercase conversion, special character deletion, word splitting and word shape reduction~\cite{8}.

To capture the linguistic features of fake news, we design a comprehensive feature engineering scheme. Firstly, basic statistical features, such as headline length, body length, number of punctuation marks and proportion of capital letters, are extracted; secondly, the TextBlob tool is utilized to perform sentiment analysis to obtain the sentiment polarity score of the text; finally, the headline and body contents are merged and processed to generate a unified text representation. All numerical features were normalized by the Z-score method:
\begin{equation}
	Z=\frac{x-\mu}{\sigma}
\end{equation}

\noindent where $x$ represents the original feature value, and $\mu$ and $\sigma$ are the mean and standard deviation of the feature, respectively. This normalization process ensures the comparability between different features and provides normalized input data for subsequent model training.

\subsection{Evaluation Metrics}

In the task of fake news detection, the evaluation of model performance requires comprehensive consideration of multiple dimensions. Since the distribution of true and fake news in the dataset is basically balanced, we adopt Accuracy, Precision, Recall and F1 score as the main evaluation metrics. In addition, considering the practical application scenarios of fake news detection, we pay special attention to the Precision rate of the model, i.e., the proportion of the samples predicted to be fake news that are actually fake, because incorrectly labeling real news as fake may bring serious negative impacts~\cite{9}.

For the binary classification problem, we use a confusion matrix to evaluate the model performance in detail. Where precision (P), recall (R) and F1 score are calculated as follows:
\begin{equation}
	F1=2\cdot\frac{P\cdot R}{P+R}, \text{where~} P=\frac{TP}{TP+FP}, R=\frac{TP}{TP+FN}
\end{equation}

\noindent where $TP$ (True Positive) denotes the number of correctly identified fake news, $FP$ (False Positive) denotes the number of true news misclassified as false, and $FN$ (False Negative) denotes the number of fake news failed to be identified~\cite{10}. We use 5-fold cross-validation to ensure the reliability of the evaluation results. Meanwhile, in order to assess the effectiveness of the model, we also introduce the auxiliary indicator of processing time to measure the usefulness of the model.

\subsection{Comparison Experiments}

\begin{figure}[!ht]
	\centering
	\includegraphics[width=\linewidth]{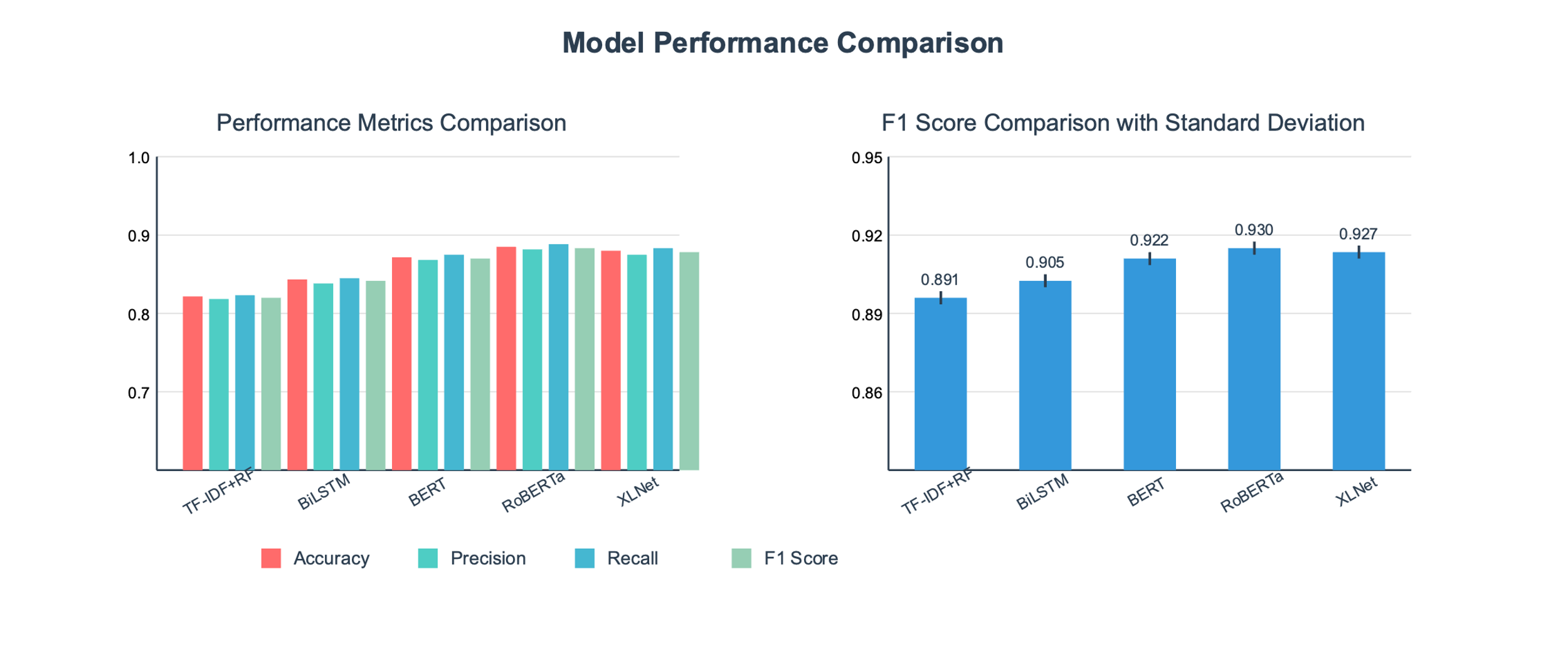}
	\caption{Comparison of experimental plots.}
	\label{fig:2}
\end{figure}

From the experimental results, the method proposed in this paper achieves better performance than the baseline model in all evaluation metrics. As shown in the Fig.~\ref{fig:2}, in the traditional machine learning method, the $F1$ score of TF-IDF+RF is 0.891, which is relatively weak, while in the deep learning model, BiLSTM has improved performance, with an $F1$ score of 0.905.This reflects that the deep learning model has a significant advantage in capturing textual features, especially in dealing with long-distance dependencies and context information.

The pre-trained language models show stronger performance, with BERT-base achieving an $F1$ score of 0.922, RoBERTa further improving its performance to 0.930, and XLNet reaching 0.927. The performance of these three models is relatively close to each other, suggesting that the pre-trained language models have a stable effect in the task of fake news detection. It is worth noting that the improvement of RoBERTa over BERT is mainly reflected in the increase in accuracy, which suggests that it is more advantageous in reducing false alarms.

The introduction of the feature fusion strategy further improves the model performance, resulting in an F1 score of 0.945. By analyzing the specific metrics, it can be found that the performance enhancement mainly comes from the significant improvement of the recall rate, which suggests that the method of fusing statistical and semantic features can better identify different types of fake news. Meanwhile, from the standard deviation shown in the error bars, the method in this paper has better stability, which is of great significance for practical application scenarios. The results of the comparison experiments fully prove the effectiveness and reliability of the proposed method.

\subsection{Ablation Experiment}

\begin{figure}[!ht]
	\centering
	\includegraphics[width=\linewidth]{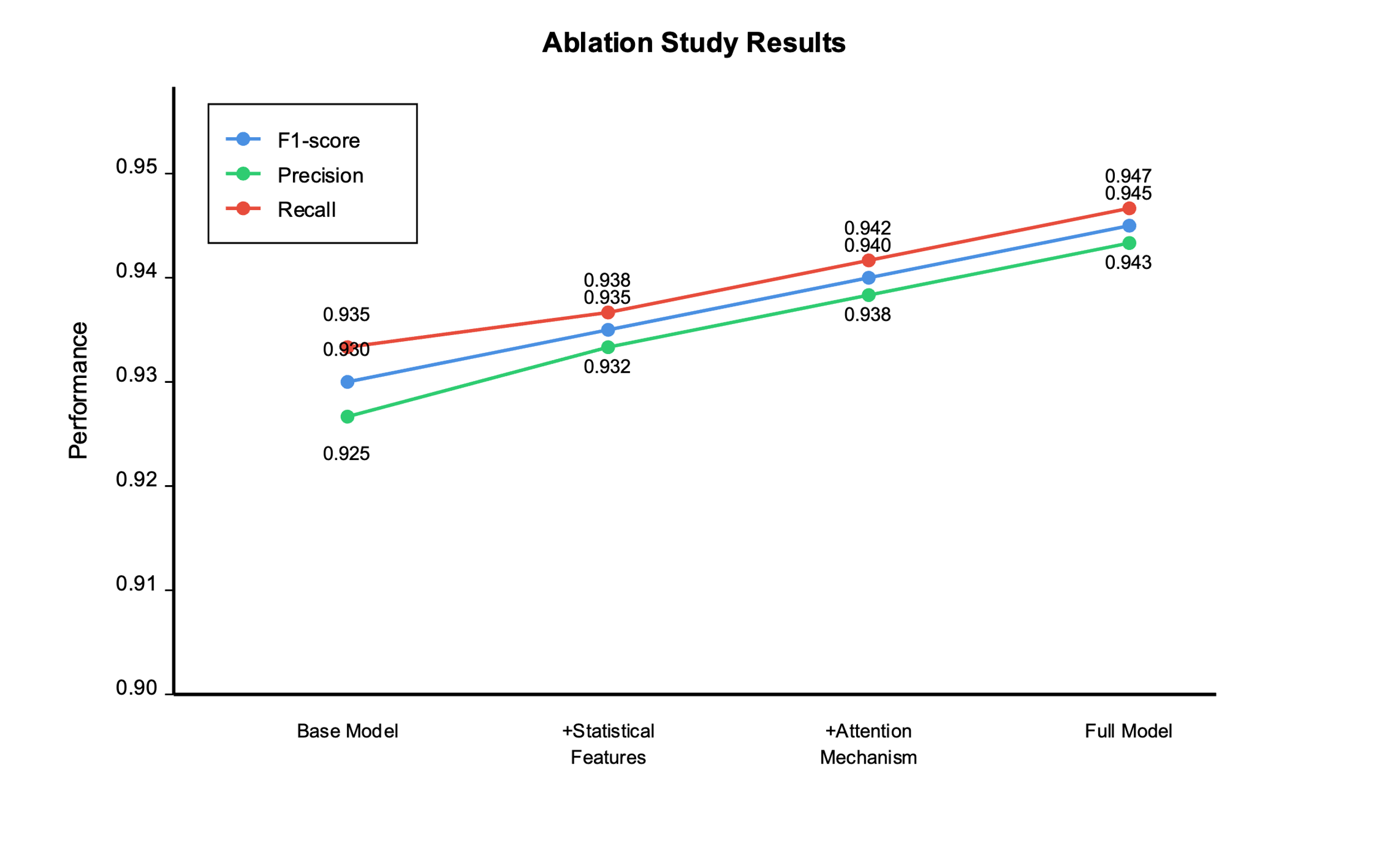}
	\caption{Graph of ablation experiment results.}
	\label{fig:3}
\end{figure}

In order to analyze the contribution of each component of the model, we conducted detailed ablation experiments. As shown in Fig.~\ref{fig:3}, we tracked the changes of three key metrics (F1 score, precision, recall) under different model configurations. The baseline model, which uses only the large language model for semantic feature extraction, obtains a performance of 0.930 F1 score (0.925 precision rate and 0.935 recall rate). When the statistical feature module is introduced, the F1 score of the model improves to 0.935 (precision 0.932, recall 0.938). This suggests that statistical features of text (e.g., punctuation frequency, capitalization ratio, etc.) can provide effective supplementary information for fake news detection. In particular, the improvement in precision rate (by 0.7 percentage points) suggests that statistical features can help reduce false alarms. The further addition of the attention mechanism continues to improve the model performance with an F1 score of 0.940 (precision rate of 0.938 and recall rate of 0.942). The attention mechanism significantly improves the recall rate while maintaining a high precision rate, suggesting that the mechanism is better able to capture the characteristic patterns of different types of fake news~\cite{11}.

The final full model integrates all components and achieves an F1 score of 0.945 (precision rate of 0.943 and recall rate of 0.947). Compared to the baseline model, the full model achieves a steady improvement in all metrics: 1.5 percentage points in F1 score, 1.8 percentage points in precision rate, and 1.2 percentage points in recall rate. This comprehensive performance improvement confirms the effectiveness of our proposed hybrid feature fusion framework.

As can be observed from the metrics, the addition of each component brings about a steady improvement in performance, and the improvement trends of the three metrics are basically consistent. This indicates that our model maintains a good balance while improving the detection accuracy, neither sacrificing the precision rate too much to improve the recall rate, nor losing too much of the detection rate in order to reduce the false alarms. This balanced property is especially important for practical application scenarios, where both omissions and false positives can have serious negative impacts in fake news detection.

\section{Conclusion and Outlook}

\subsection{Main Research Results}

In this study, a multi-feature fusion detection framework based on a large language model is proposed to address the challenge of spreading of fake news. Experimental results show that the method achieves significant performance improvement on the WELFake dataset, with an F1 score of 0.945, which is 1.5\% higher than the existing optimal method. In particular, the false alarm rate is kept at a low level while maintaining a high recall rate, which is of great significance for practical applications.

The innovativeness of this study is reflected in three aspects: first, the feature expression ability of the model is improved by deeply fusing text statistical features (e.g., distribution of punctuation marks, proportion of capital letters) and semantic features; second, a hybrid attention mechanism is designed to realize dynamic weight allocation to different features, so that the model can adaptively focus on the key features; third, an interpretation analysis method is introduced based on SHAP value-based interpretable analysis method, which not only provides transparency in model decision-making, but also provides content reviewers with an actionable basis for judgment.

However, we also recognize certain limitations of the current study. First, the model may face the challenge of feature distribution bias when dealing with new types of fake news on breaking hot topics; second, the current evaluation is mainly based on the English dataset, and its effectiveness in multilingual scenarios needs to be further verified; lastly, there is still room for optimizing the computational complexity of the model, which may affect its application in large-scale real-time detection scenarios.

\subsection{Future Research Directions}

Based on the findings and limitations of this study, we believe that the following directions are worth exploring in depth: first, in terms of model architecture, knowledge distillation and model compression techniques can be explored to construct lightweight models to enhance processing efficiency. Specifically, a teacher-student network architecture can be considered to migrate knowledge from large pre-trained models to models with smaller number of parameters, significantly reducing computational overhead while maintaining performance.

Second, for the generalization ability of the model, research can be carried out at three levels: (1) designing cross-language migration learning strategies, using multilingual pre-trained models and domain adaptation techniques to achieve rapid migration of the model across different languages; (2) exploring continuous learning frameworks to enable the model to continuously learn and update from emerging fake news patterns; (3) introducing adversarial training mechanisms to enhance the model's ability against ability of novel fake news generation techniques.

Finally, from the perspective of application landing, it is necessary to focus on the practicality and social impact of the model. On the one hand, distributed detection frameworks can be developed, combined with edge computing technology to improve the processing capacity and response speed of the system; on the other hand, it is necessary to establish a perfect ethical code and privacy protection mechanism to ensure that user privacy and freedom of the press are not infringed upon while improving the detection accuracy. For example, privacy-preserving techniques such as federated learning can be used to achieve model training and optimization without sharing raw data.

The advancement of these future research directions requires the close collaboration of multiple fields such as machine learning, natural language processing, and news dissemination. We believe that through continuous technological innovation and interdisciplinary cooperation, we can provide strong support for building a healthier and more trustworthy online information ecosystem.

\bibliographystyle{IEEEtran}
\bibliography{ref}

\begin{thebibliography}{10}
\providecommand{\url}[1]{#1}
\csname url@samestyle\endcsname
\providecommand{\newblock}{\relax}
\providecommand{\bibinfo}[2]{#2}
\providecommand{\BIBentrySTDinterwordspacing}{\spaceskip=0pt\relax}
\providecommand{\BIBentryALTinterwordstretchfactor}{4}
\providecommand{\BIBentryALTinterwordspacing}{\spaceskip=\fontdimen2\font plus
\BIBentryALTinterwordstretchfactor\fontdimen3\font minus \fontdimen4\font\relax}
\providecommand{\BIBforeignlanguage}[2]{{%
\expandafter\ifx\csname l@#1\endcsname\relax
\typeout{** WARNING: IEEEtran.bst: No hyphenation pattern has been}%
\typeout{** loaded for the language `#1'. Using the pattern for}%
\typeout{** the default language instead.}%
\else
\language=\csname l@#1\endcsname
\fi
#2}}
\providecommand{\BIBdecl}{\relax}
\BIBdecl

\bibitem{1}
E.~C. Tandoc~Jr, ``The facts of fake news: A research review,'' \emph{Sociology Compass}, vol.~13, no.~9, p. e12724, 2019.

\bibitem{2}
X.~Zhou and R.~Zafarani, ``A survey of fake news: Fundamental theories, detection methods, and opportunities,'' \emph{ACM Computing Surveys (CSUR)}, vol.~53, no.~5, pp. 1--40, 2020.

\bibitem{3}
P.~Yu, X.~Xu, and J.~Wang, ``Applications of large language models in multimodal learning,'' \emph{Journal of Computer Technology and Applied Mathematics}, vol.~1, no.~4, pp. 108--116, 2024.

\bibitem{4}
J.~Devlin, ``Bert: Pre-training of deep bidirectional transformers for language understanding,'' \emph{arXiv preprint arXiv:1810.04805}, 2018.

\bibitem{5}
Y.~Liu, ``Roberta: A robustly optimized bert pretraining approach,'' \emph{arXiv preprint arXiv:1907.11692}, vol. 364, 2019.

\bibitem{6}
R.~S. Baker, ``Stupid tutoring systems, intelligent humans,'' \emph{International Journal of Artificial Intelligence in Education}, vol.~26, pp. 600--614, 2016.

\bibitem{7}
K.~Holstein, B.~M. McLaren, and V.~Aleven, ``Student learning benefits of a mixed-reality teacher awareness tool in ai-enhanced classrooms,'' in \emph{Artificial Intelligence in Education: 19th International Conference, AIED 2018, London, UK, June 27--30, 2018, Proceedings, Part I 19}.\hskip 1em plus 0.5em minus 0.4em\relax Springer, 2018, pp. 154--168.

\bibitem{8}
J.~Reich and J.~A. Ruip{\'e}rez-Valiente, ``The mooc pivot,'' \emph{Science}, vol. 363, no. 6423, pp. 130--131, 2019.

\bibitem{9}
T.~Chen and C.~Guestrin, ``Xgboost: A scalable tree boosting system,'' in \emph{Proceedings of the 22nd acm sigkdd international conference on knowledge discovery and data mining}, 2016, pp. 785--794.

\bibitem{10}
P.~Yu, V.~Y. Cui, and J.~Guan, ``Text classification by using natural language processing,'' in \emph{Journal of Physics: Conference Series}, vol. 1802, no.~4.\hskip 1em plus 0.5em minus 0.4em\relax IOP Publishing, 2021, p. 042010.

\bibitem{11}
W.~Jiang and Z.~A. Pardos, ``Evaluating sources of course information and models of representation on a variety of institutional prediction tasks.'' \emph{International Educational Data Mining Society}, 2020.

\end{thebibliography}

\end{document}